# Physics-Informed Deep Learning: A Promising Technique for System Reliability Assessment

Taotao Zhou[1*], Enrique Lopez Droguett[2,3], Ali Mosleh[3]

[1] *Center for Risk and Reliability, University of Maryland, College Park, USA*
[2] *Department of Civil and Environmental Engineering, University of California, Los Angeles, USA*
[3] *The Garrick Institute for the Risk Sciences, University of California, Los Angeles, USA*

***ABSTRACT:***
Considerable research has been devoted to deep learning-based predictive models for system prognostics and health management in the reliability and safety community. However, there is limited study on the utilization of deep learning for system reliability assessment. This paper aims to bridge this gap and explore this new interface between deep learning and system reliability assessment by exploiting the recent advances of physics-informed deep learning. Particularly, we present an approach to frame system reliability assessment in the context of physics-informed deep learning and discuss the potential value of physics-informed generative adversarial networks for the uncertainty quantification and measurement data incorporation in system reliability assessment. The proposed approach is demonstrated by three numerical examples involving a dual-processor computing system. The results indicate the potential value of physics-informed deep learning to alleviate computational challenges and combine measurement data and mathematical models for system reliability assessment.

***KEYWORDS:*** physics-informed deep learning, reliability assessment, generative adversarial networks, uncertainty quantification.

---

[*] Corresponding author.
E-mail address: taotao.zhou@outlook.com (Taotao Zhou).

# 1. INTRODUCTION

Deep learning has been an emerging approach to handle multi-dimensional sensor data without requiring many manual feature engineering efforts [1]. This draws increasing attention from the reliability and safety community to developing the deep learning-based predictive framework for prognostics and health management (PHM), which has been comprehensively documented in some review articles [2, 3, 4]. From a reliability perspective, these studies address some particular challenges encountered in the reliability context: (i) develop probabilistic deep learning models that consider epistemic and aleatory uncertainties in support of the decision-making process [5, 6, 7]; (ii) develop deep domain adaptation and generalization models that address the challenges of varying working conditions; (iii) develop hybrid deep learning models that exploit both physics knowledge and data [8]; (iv) study the robustness of deep learning-based PHM models under adversarial attacks [9, 10]; (v) address the imbalanced dataset due to the scarcity of fault observations [11]; (vi) integrate deep learning-based models into the conventional probabilistic risk assessment (PRA) that incorporates the specific systems' dynamic evolutions [12].

All the above research mainly focuses on deep learning-based PHM using the sensor data that captures the system's health conditions. There is little literature available to explore the value of deep learning in the system reliability assessment. Typically, the system reliability evolution is represented by mathematical models. For instance, partial differential equations describe the underlying failure mechanisms [13, 14], Markov process characterizes the state transitions using a set of ordinary differential equations [15]. These problems are usually difficult to solve analytically, and hence numerical methods are commonly used, such as differential equation solver and Monte Carlo simulation. These numerical methods are computational expensive especially when uncertainty and sensitivity analysis is required for safety-critical applications. Moreover, those methods cannot integrate the measurement data collected through inspection and maintenance activities.

Recent advances in scientific machine learning provide a new lens to solve the aforementioned problems [16]. The physics-informed neural networks (PINNs) have been well-recognized since the pioneering work of Raissi et al. [17]. The essential idea is to use neural networks as universal approximators of the desired solution and then constrain the training process by designing the loss function according to the domain-specific knowledge, such as the physics model described by partial differential equations. Most recent research focuses on enhancing PINNs by using different deep learning architectures [18, 19, 20] with applications to biophysics, geophysics, and engineering sciences [21, 22, 23, 24, 25]. The interested readers can find a comprehensive review of the progress of physics-informed deep learning with diverse applications in Karniadakis et al. [26].

Physics-informed deep learning is still in an early stage of development and needs to be well configured given the specific problem. One of the main concerns is to improve PINNs for uncertainty quantification to achieve a robust and reliable prediction. Most notably, physics-informed generative adversarial networks (PIGANs) have been proposed and are still under development to probabilistically leverage observations and the underlying physics model in solving both forward and inverse problems. The studies of PIGANs mainly use a similar framework and vary depending on two factors: integrate domain knowledge into either generator or discriminator [27]; adopt different types of GANs to improve their training stability [18, 19, 28, 29, 30]. The choice of PIGANs would vary depending on the scale of specific problems and the computational resources available. It is worthwhile noting that the above research also shows great value to address the challenges of scarcity of measurements in system reliability assessment. Particularly, the following applications adopted the PIGANs proposed by Yang and Perdikaris [18], which integrates domain knowledge into the generator and follows the training scheme developed by Li et al. [31].

This paper presents a novel perspective of system reliability assessment by leveraging the advance of physics-informed deep learning. The main objective is to make a connection between deep learning and system reliability assessment and to further demonstrate the important value of deep learning to benefit system reliability assessment. Our contributions are twofold:

1) We present an approach to frame system reliability assessment as a problem of deep learning. The essential is to encode the system property into the network configuration and training based on the mathematical model governing system reliability evolutions. Particularly, approximate the solution to reliability assessment with a neural network and induce another neural network to obtain the derivatives of system state probability by using automatic differentiate techniques. The outputs of the two neural networks are utilized to construct a composite loss function, and then gradient-based optimization algorithms are employed to learn the solution to system reliability assessment. It is worthwhile noting that this provides a continuous solution to the system reliability assessment because of the universal approximation theorem [32]. This enables one to assess system reliability at any given time instant.
2) To highlight the potential value of physics-informed deep learning, we discuss a PIGANs-based approach for uncertainty quantitation and incorporating measurement data in system reliability assessment. The essential is to formulate a deep probabilistic setting by an adversarial game between the data constraints and the mathematical model describing system reliability evolution. This provides a new perspective on combining measurement data with the underlying mathematical model. This is particularly valuable for the safety-critical sectors with a small number of measurements, such as passive structures in nuclear power plants. Moreover, the proposed approach has superior efficiency to alleviate the computational challenges in uncertainty quantification that is important for the reliability and safety community.

The proposed approach is demonstrated using a dual-processor computing system with performance degradation. Three examples are formulated and discussed to validate the proposed approach: (i) the system starts with a perfect condition and degrades over mission time, which is validated by comparison with the differential equation solver and the Monte Carlo simulation; (ii) the system starts with either a perfect condition or a degraded state. The uncertainty is modeled by the Bernoulli distribution, the epistemic uncertainty of which is modeled by the Beta distribution. The results are validated by comparison with the Monte Carlo simulation; (iii) the system starts with a perfect condition given synthetic measurement data available to reflect the system's condition during a service life span. This example is heuristically demonstrated by two simulated systems with either better or worse performance as compared to the baseline case in the first example. A heuristic demonstration is presented because measurement data cannot be properly incorporated using the current methods for system reliability assessment. Overall, the results validate the effectiveness of the proposed approach for system reliability assessment and show the superiority of the proposed approach in computational efficiency.

This remaining paper is structured as follows. Section 2 summarizes the problem formulation and background of system reliability assessment. Section 3 presents the deep learning-based approach for system reliability assessment. Section 4 demonstrates the proposed model using three numerical examples involving a dual-processor computing system. Section 5 discusses the conclusions and future directions.

## 2. PROBLEM FORMULATION AND BACKGROUND

Conventionally, the component or system performance is modeled as a binary state that is either fully functioning or complete failure. However, this is often not the truth, especially, where an intermediate state always manifests between fully functioning and complete failure. Therefore, it is a natural fit to characterize the component or system states by a range of discrete levels, often known as a multi-state model [33]. Typically, the multi-state model is mathematically described by a continuous-time Markov or semi-Markov process in reliability applications. We limit the scope of this paper to the continuous-time Markov process.

Suppose a system performance is characterized by a finite number of states $S = \{0, 1, \ldots, j, \ldots, M\}$. The system dynamics are reflected by the transitions across states at each time instant $t$, which is parameterized by the transition rate $\lambda_{i,j}(t)$ from state $i$ to state $j$. Note that the state transitions are often time-dependent, due to performance deterioration and maintenance interventions. Hence, denote the corresponding transition rate matrix $Q(t)$ at a time instant $t$ as below, where $\lambda_i(t) = \sum_{j \in S, j \neq i} \lambda_{i,j}(t)$.

$$Q(t) = \begin{bmatrix} -\lambda_0(t) & \lambda_{0,1}(t) & \ldots & \lambda_{0,j}(t) & \ldots & \lambda_{0,M}(t) \\ \lambda_{1,0}(t) & -\lambda_1(t) & \ldots & \lambda_{1,j}(t) & \ldots & \lambda_{1,M}(t) \\ \ldots & \ldots & \ldots & \ldots & \ldots & \ldots \\ \lambda_{j,0}(t) & \lambda_{j,1}(t) & \ldots & -\lambda_j(t) & \ldots & \lambda_{j,M}(t) \\ \ldots & \ldots & \ldots & \ldots & \ldots & \ldots \\ \lambda_{M,0}(t) & \lambda_{M,1}(t) & \ldots & \lambda_{M,j}(t) & \ldots & -\lambda_M(t) \end{bmatrix} \quad (1)$$

The system state at each time instant $t$ is represented by a probability vector, that is $p(t) = \{p_0(t), p_1(t), \ldots, p_j(t), \ldots, p_M(t)\}$, where $p_j(t)$ is the probability that the system is in state $j$ at time instant $t$, and $\sum_{j=1}^{M} p_j(t) = 1$. The system state probability can be derived according to the forward Kolmogorov equations, which consists of a set of differential equations parameterized by the transition rate matrix and state probability vector in Equation (2). Then the system reliability can be determined by aggregating the state probability where the system is considered as functioning.

$$p'(t) = p(t)Q(t) \quad (2)$$
$$p(t = 0) = s_0 \quad (3)$$

where $p'(t)$ is the derivative of $p(t)$ for system's operational time $t$, and $p(t = 0)$ represents the system state at time instant 0, referred to as the initial condition $s_0$.

# 3. SYSTEM RELIABILITY ASSESSMENT USING DEEP LEARNING

This section discusses the deep learning-based approach for system reliability assessment in continuous time. Section 3.1 discusses the approach to frame a deep learning problem to assess system reliability. Section 3.2 discusses a PIGANs-based approach for uncertainty quantification and measurement data incorporation in system reliability assessment.

## 3.1. Frame System Reliability Assessment in Deep Learning Context

This section focuses on the connection between deep learning and system reliability assessment. The essential objective is to learn a continuous latent function as the solution to system reliability considering the possible state transitions and the initial condition. In particular, a neural network is utilized to approximate $p(t)$ which acts as a prior on the unknown reliability solution. According to the universal approximation theorem, this leads to a continuous solution that enables one to assess the system reliability at any time instant up to mission time. As illustrated in Figure 1, the system property is encoded into the network configuration and training, which are discussed in the following.

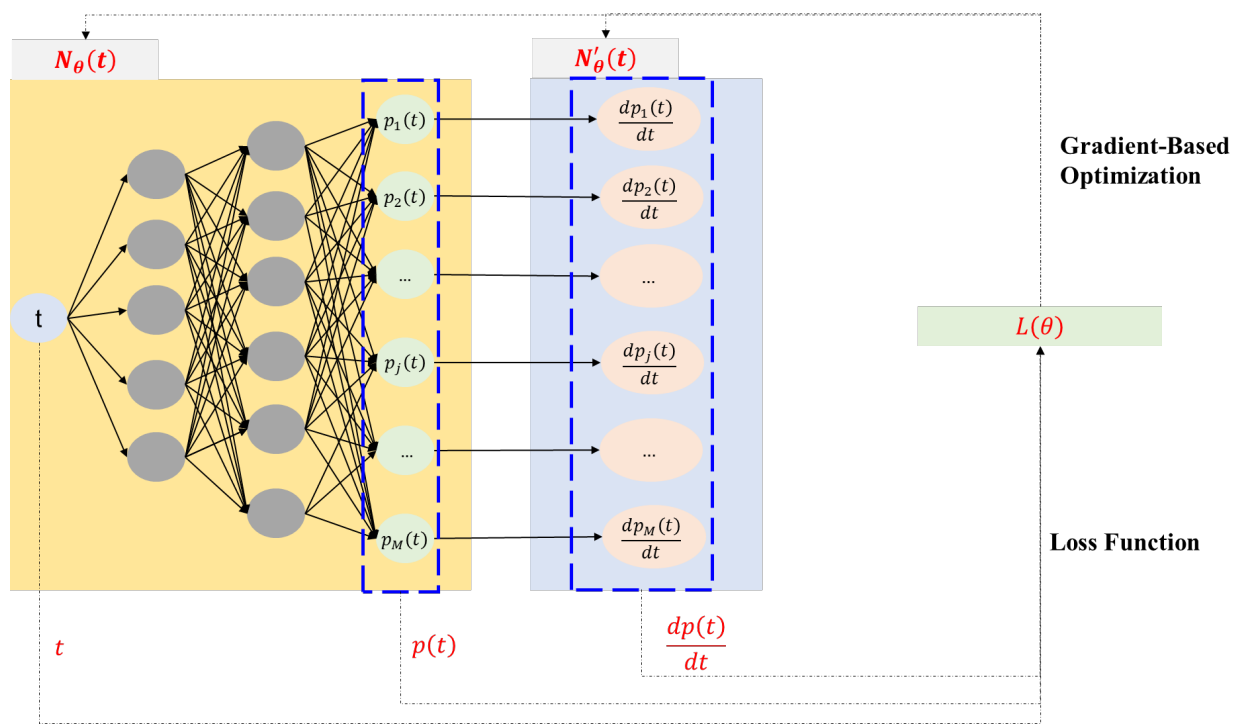

Figure 1. Frame system reliability assessment as a problem of deep learning.

In the network configuration, there are two neural networks with shared parameters that approximate the system state probability and obtain their derivative regarding the system's operational time. In other words, the time dependency is encoded by the state probability and its derivative at any time instant. These two networks are configured as below:

- Utilize a neural network $N_\theta(t)$ as the surrogate for the reliability estimates $p(t)$, where $N_\theta(t)$ denotes a neural network parametrized by $\theta$ and the network input is global time $t$. The number of neurons in the output layers needs to match the number of system states. Then using the SoftMax function as the activation function in the output layer would provide the probability regarding each system state. This implicitly satisfies the constraints of probability value regarding each state in the range [0,1].
- Establish an induced neural network $N'_\theta(t)$ to obtain the derivative of system state probability $p'(t)$. Particularly, $N'_\theta(t)$ is an induced neural network based on $N_\theta(t)$ using automatic differentiation.

The network training process needs to be constrained to satisfy the system initial condition and system state transition model as reflected in Equations (2) and (3). Therefore, a composite loss function can be constructed in Equation (4), by combining two residual terms.

$$L(\theta) = (N_\theta(t=0) - s_0)^2 + \lambda \left[\frac{1}{N_r}\sum_{i=1}^{N_r}\left(N_\theta(t_i) \cdot Q(t_i) - \frac{dN_\theta(t_i)}{dt_i}\right)^2\right] \quad (4)$$

where the first term enforces the neural network in agreement with the system's initial condition as given by Equation (3); the second term enforces training process consistent with the system state transition as expressed in Equation (2) by penalizing at $N_r$ collocation points; $\lambda$ is a weighting factor to balance those loss terms. This is then formulated as a minimization problem to train the neural networks via gradient-based optimization algorithms.

$$\min_\theta L(\theta) \quad (5)$$

The optimal parameters would be learned to parameterize the solutions to system reliability assessment. Then one can assess the system state probability at any time instant in Equation (6), where $p_j(t)$ denotes the $j^{th}$ state probability at time $t$. As shown in Equation (7), the system reliability $R(t)$ can be obtained by summing the probability regarding a set of state indexes $U$, where the system works reliably.

$$p_j(t) \approx N_\theta(t)[j] \quad (6)$$

$$R(t) \approx \sum_{j \in U} N_\theta(t)[j] \quad (7)$$

## 3.2. Physics-Informed Generative Adversarial Networks (GANs) For System Reliability Assessment

This section further discusses how deep learning can benefit system reliability assessment. Particularly, we discuss PIGANs based approach to integrate data constraints in system reliability assessment. The idea is to formulate a probabilistic setting to learn the probabilistic distributions of the system reliability that, in turn, results in a generative model capable to produce synthetic data, which is consistent with data constraints and underlying mathematical model describing system reliability evolution. The data constraints would be imposed by measurement data observed through the system's lifetime and the system's initial condition. As displayed in Figure 2, the key is to encode both system property and data constraints into the network configuration and training, by formulating an adversarial game between a generator and a discriminator. Without loss of generality, the reasoning behind the network design and training is discussed based on standard GANs conditional on the system's operational time $t$.

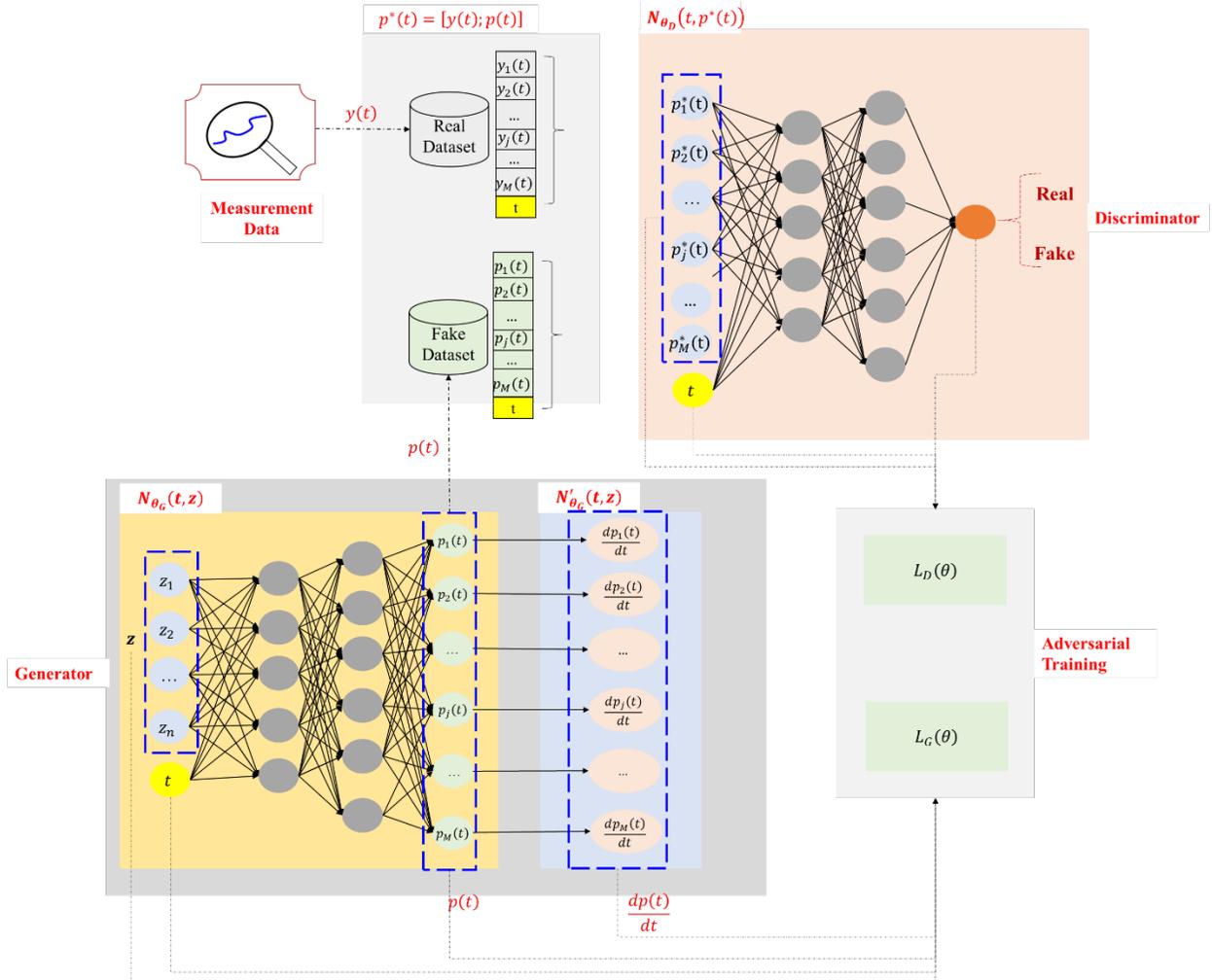

Figure 2. A physics-informed generative adversarial network (GANs) based approach for system reliability assessment considering measurement data.

The generator $N_{\theta_G}(t, Z)$ is designed to produce fake datasets given random noise vector $Z$ and time $t$. $Z$ is a collection of random latent variables with a multivariate Gaussian distribution and is employed to construct probabilistic representations for the system state probability. The fake data needs to approximately satisfy the underlying state transitions governed by Equation (2). This generator is designed in the same manner as Section 3.1, with an additional loss term to constrain the learning process to be consistent with the measurement data. The real measurement data is denoted by $[t_j, y(t_j)]$, where $j = 1,2 \dots N_d$. The system's initial condition can also be treated as one type of measurement data. Then, we can combine it with the measurement data as $[t_k, \; y(t_k)]$, where $k = 0,1,2 \dots N_d$ and $y(t_0 = 0) = s_0$. The generator loss function is shown in Equation (8). The first term fools the discriminator to mark the fake dataset as the real ones; the second term constrains the generated fake data to be consistent with the measurement data; the third term imposes the constraints according to the domain knowledge.

$$L_G(\theta_G) = \frac{1}{N_d + 1} \sum_{k=0}^{N_d} \log\left(1 - N_{\theta_D}\left(t_k, N_{\theta_G}(t_k, z_k)\right)\right)$$
$$+ \frac{1}{N_d + 1} \sum_{k=0}^{N_d} \left(y(t_k) - N_{\theta_G}(t_k, z_k)\right)^2 \qquad (8)$$
$$+ \lambda \left[\frac{1}{N_r} \sum_{i=1}^{N_r} \left(N_{\theta_G}(t_i, z_i) \cdot Q(t_i) - \frac{dN_{\theta_G}(t_i, z_i)}{dt_i}\right)^2\right]$$

The discriminator $N_{\theta_D}(t, u)$ is designed to distinguish between fake datasets produced by the generative model and real datasets collected from measurements. There is one neuron in the output layer with the sigmoid activation function. This constrains the generator to produce a dataset matching the real dataset. The discriminator loss function is shown in Equation (9). The first term is maximized to correctly classify the real measurement data, and the second term is maximized to correctly detect the fake dataset produced by the generator.

$$L_D(\theta_D) = \frac{1}{N_d + 1} \sum_{k=1}^{N_d} \log\left(N_{\theta_D}(t_k, y(t_k))\right)$$
$$+ \frac{1}{N_d + 1} \sum_{k=0}^{N_d} \log\left(1 - N_{\theta_D}\left(t_k, N_{\theta_G}(t_k, z_k)\right)\right) \qquad (9)$$

Thus, the two competing loss functions are used: $L_G(\theta_G)$ and $L_D(\theta_D)$, for the generator and discriminator, respectively. Then, we derive an adversarial training rule for updating the unknown model parameters contained in the vectors $\theta_D, \theta_G$. This leads to an adversarial game for training the PIGANs by alternating the optimization of the two objectives in Equations (10) and (11).

$$\max_{\theta_D} L_D(\theta_D) \tag{10}$$

$$\min_{\theta_G} L_G(\theta_G) \tag{11}$$

Upon the PIGANs model is successfully trained, the generator $N_{\theta_G}(t, z)$ can be utilized to simulate the system reliability considering measurement data and uncertainty. Particularly, the system reliability assessment is accomplished by drawing $N_s$ samples through stochastic forward passes of the generator. A point estimate of $j^{th}$ state probability is determined by computing the mean of the predictions regarding each sample in Equation (12). The uncertainty of $j^{th}$ state probability is characterized by intervals with two-standard deviation in Equation (13).

$$\overline{p_j(t)} \approx \frac{1}{N_s} \sum_{n=1}^{N_s} N_{\theta_G}(t, z_n)[j] \tag{12}$$

$$\sigma_j(t) \approx \sqrt{\frac{1}{N_s - 1} \sum_{n=1}^{N_s} \left[N_{\theta_G}(t, z_n)[j] - \overline{p_j(t)}\right]^2} \tag{13}$$

Assume the system works reliably in a set of state indexes $U$. It is straightforward to compute the point estimate and uncertainty of system reliability by aggregating the state probability within the set U as follows:

$$\overline{R(t)} \approx \frac{1}{N_s} \sum_{n=1}^{N_s} \sum_{j \in U} N_{\theta_G}(t, z_n)[j] \tag{14}$$

$$\sigma_R(t) \approx \sqrt{\frac{1}{N_s - 1} \sum_{n=1}^{N_s} \left[\sum_{j \in U} N_{\theta_G}(t, z_n)[j] - \overline{R(t)}\right]^2} \tag{15}$$

The PIGANs-based approach has twofold advantages. First, the generator can be used as a surrogate model for conventional Monte Carlo simulation. This would be more efficient for a highly reliable system that is often computationally expensive and requires a large number of samples in the conventional Monte Carlo simulation. Second, the proposed approach accounts for both measurement data and the mathematical model, thus, in turn, the system reliability evolution simulated by the generator is also informed by the measurement data. This offers a unique advantage against the current methods that cannot consider measurement data. These two advantages are experimentally demonstrated in the following numerical examples.

## 4. NUMERICAL EXAMPLES

This section demonstrates the deep learning approach for system reliability assessment using three numerical examples involving a four-state system. Section 4.1 provides a brief description of problem formulation. Section 4.2 discusses the results and the model performance assessment. The proposed approach was developed based on Python v3.6 [34], and TensorFlow v1.13 [35] using a desktop with Intel Core i7-6700 CPU and 32 GB DDR4 RAM. The differential equation solver was implemented using Matlab [36]. The Monte Carlo simulation was also implemented in Python v3.6 [34].

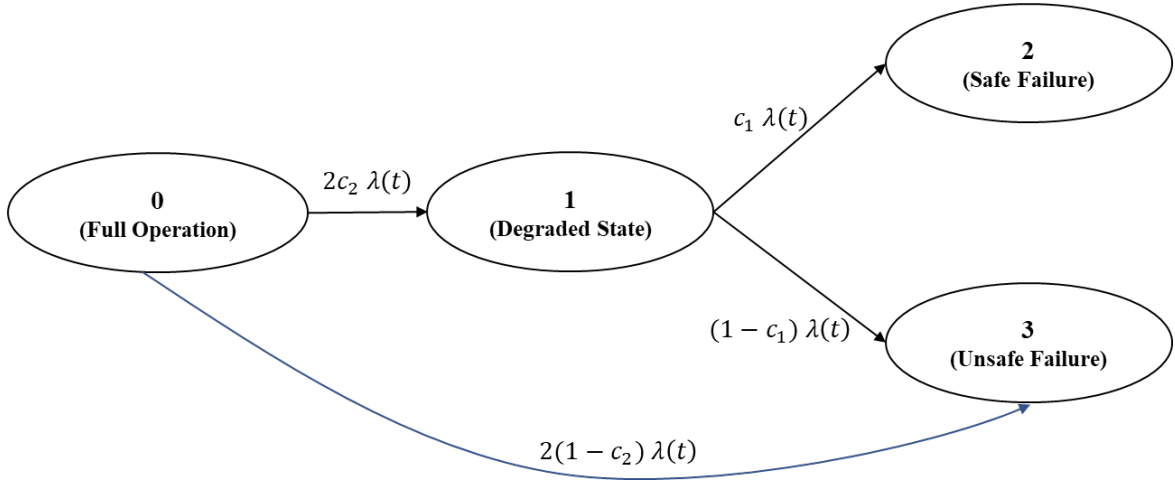

Figure 3. A state transition diagram to describe the performance deterioration of a dual-processor computing system.

### 4.1. Problem Description

Consider a safety model of a dual-processor computing system that degrades through 4 possible states {0, 1, 2, 3} as taken from Rindos et al. [37]. Figure 3 shows the state transition diagram describing the possible transition across states. With the system state increasing from 0 to 3, the system continuously degrades until a safe or unsafe failure. The system is considered reliable in states 0 and 1, so the system reliability can be calculated by summing the probability of these two states. The definition of each state is presented below.

- State 0: the system functions in full capacity with two processors.
- State 1: the system works in a degraded mode given any of the two processors fails, which can be successfully detected (i.e., probability $c_2$).
- State 2: the system is operated in a degraded state, and the other processor failure leads to safe shutdown (i.e., probability $c_1$).

- State 3: the system fails unsafely due to two scenarios: any of the two processors fails but is not detected (i.e., probability $1 - c_2$); the failure of the other processor leads the system to an unsafe state (i.e., probability $1 - c_1$) when the system is operated in a degraded state.

The transition across states follows the Weibull distribution, and the transition rates are denoted by $\lambda(t) = \lambda_0 \alpha t^{\alpha-1}$, where $t$ is the system's operational time. This results in a Non-Homogenous Continuous-Time Markov system where the transition rates depend on the system's operational time. In this paper, we set the parameters as the same as Rindos et al. [37], that is $c_2 = 0.9$, $c_1 = 0.9$, $\lambda_0 = 0.01$ and $\alpha = 2.0$. The corresponding transition rate matrix is shown in Equation (16).

$$\boldsymbol{Q}(t) = \begin{bmatrix} -0.04 \cdot t & 0.036 \cdot t & 0 & 0.004 \cdot t \\ 0 & -0.02 \cdot t & 0.018 \cdot t & 0.002 \cdot t \\ 0 & 0 & 0 & 0 \\ 0 & 0 & 0 & 0 \end{bmatrix} \tag{16}$$

To demonstrate our proposed approach, three example problems are formulated with the detailed setup below:

1) Suppose the system starts with a perfect working condition (i.e., state 0) and degrades over a service life span. The results are validated by a comparative study with the ordinary differential equation solver and the Monte Carlo simulation, respectively.
2) Suppose the system starts with a perfect working condition (i.e., state 0) or degraded state (i.e., state 1), which follows the Bernoulli distribution. This scenario is subject to large uncertainty due to the cause of manufacturing defects and installation in the field. Hence, we use the Beta distribution to model the epistemic uncertainty for the Bernoulli distribution. The results are validated by comparison with the Monte Carlo simulation.
3) The third example intends to demonstrate how the deep learning approach can incorporate the measurement data into the mathematical model describing the underlying state transitions. Particularly, we follow the same assumption that the system starts with a perfect working condition; generate some synthetic measurement data by using the results in the first example as a baseline; heuristically validate the results by discussing the impacts of measurements on the system behavior.

## 4.2. Results and Discussions

This section discusses the results of the three examples. For validation purposes, all the following discussions are based on the system reliability with a certain time step (i.e., 1-time unit) equally up to mission time 30.

*4.2.1. Example 1*

The system reliability is assessed using the proposed approach in Section 3.1. The neural network consists of 2 hidden layers, each of which has 50 neurons and uses the Tanh activation function. There are four neurons in the output layer with the SoftMax activation function. The output of each neuron corresponds to the probability of each system state, respectively. There are 40 collocation points, which are generated linearly spaced within the range [0, 30]. The network is trained using the Adam optimization algorithm and the number of iterations is $2 \times 10^4$. An exponential decaying learning rate is applied with the starting learning rate as $1 \times 10^{-3}$, the decay rate as 0.9 and the decay step as 1000. The weighting factor $\lambda$ is set equal to 1. On the other hand, baseline results are obtained by using the differential equation solver particularly the Runge-Kutta method and the Monte Carlo simulation with $1 \times 10^5$ iterations. Figure 4 summarizes the mean value of each system state probability. The results of the proposed approach are close to the baseline results, which indicates the good performance of our proposed approach.

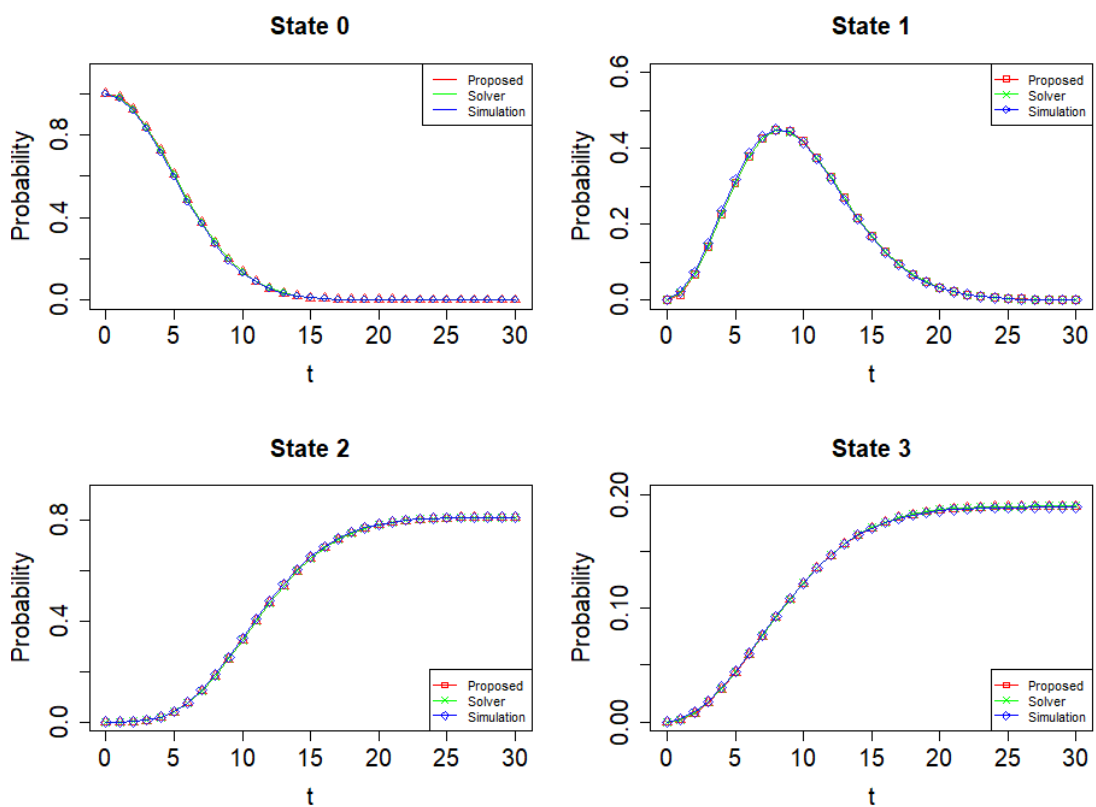

Figure 4. The results of system state probability using the proposed approach, the differential equation solver, and the Monte Carlo simulation.

To further evaluate the consistency of the results between the proposed approach and the differential equation solver, we run the proposed approach for 60 replications considering the random effects through training and testing. The consistency between the results is measured using the root mean square error (RMSE) in Equation (17), where $p_j^*(t)$ is the probability of state $j$ at time $t$ given by the differential equation solver, $p_j^i(t)$ is the state probability of state $j$ at time $t$ obtained in the $i^{th}$ replication of the proposed approach, the total number of replications is $N$, $RMSE_j(t)$ is the RMSE of $j^{th}$ state probability between the proposed approach and differential solver at time $t$.

$$RMSE_j(t) = \sqrt{\frac{1}{N}\sum_{i=1}^{N}[p_j^i(t) - p_j^*(t)]^2} \qquad (17)$$

The distributions of $RMSE_j(t)$ up to mission time are summarized in Figure 5. The overall variability of the RMSE is further illustrated by its value of 5% quantile, median, mean, and 95% quantile. The RMSE remains relatively small and indicates the effectiveness of our proposed method to achieve a satisfactory assessment of system reliability.

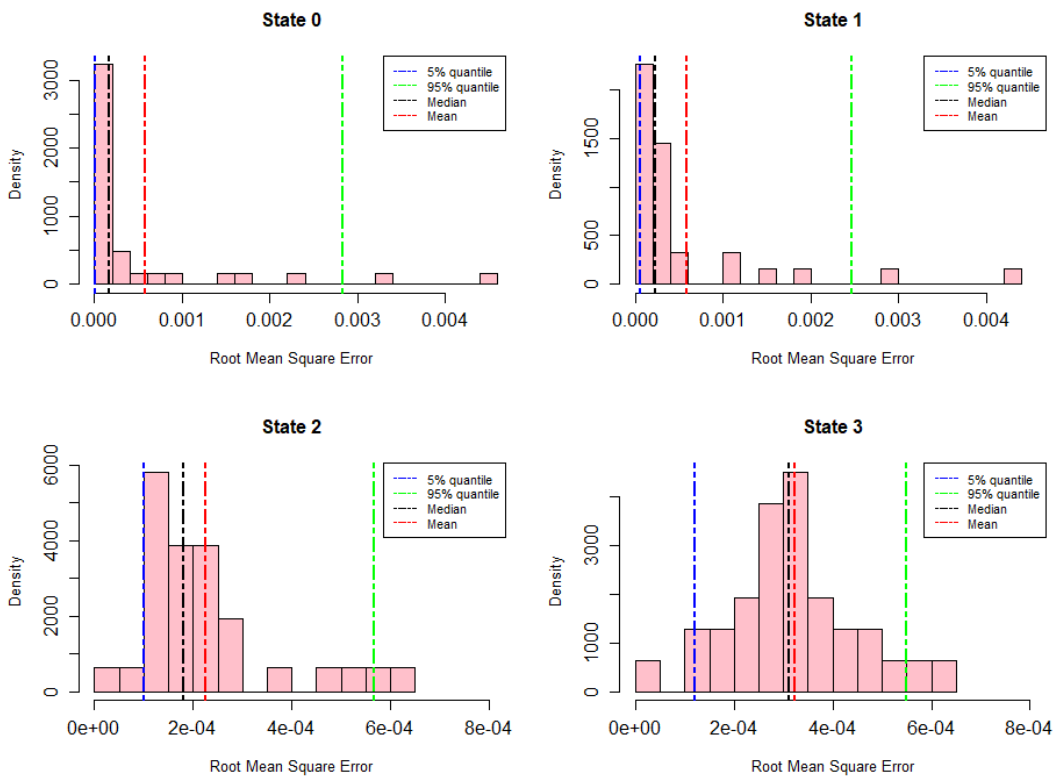

Figure 5. The root mean square error (RMSE) of the results between the proposed approach and the differential equation solver.

The proposed approach is also validated by comparison with the Monte Carlo simulation. Considering the random effects of both methods, we run the proposed approach and the Monte Carlo simulation for 60 replications, respectively. The predictive uncertainty at each time instant is characterized by the mean and standard deviation of the corresponding realizations. The prediction of $j^{th}$ state probability at time $t$ is denoted: by mean $p_j(t)$ and standard deviation $\sigma_j(t)$ using the proposed approach; by mean $p'_j(t)$ and standard deviation $\sigma'_j(t)$ using the Monte Carlo simulation. Then evaluate the consistency of results between the proposed approach and the Monte Carlo simulation by measuring their absolute difference and composite standard deviation, as represented by $\Delta p_j(t)$ and $\Delta \sigma_j(t)$ in Equations (18) and (19), representatively.

$$\Delta p_j(t) = |p_j(t) - p'_j(t)| \tag{18}$$

$$\Delta \sigma_j(t) = \sqrt{\sigma_j(t)^2 + \sigma'_j(t)^2} \tag{19}$$

Figures 6 and 7 show the distribution of $\Delta p_j(t)$ and $\Delta \sigma_j(t)$ up to mission time. The values of absolute difference and composite standard deviation remain relatively small, which shows that the performance of the proposed approach is as comparable as the Monte Carlo simulation. Furthermore, it is worthwhile noting that the proposed approach only takes 17.9 seconds for a replicate, while a replicate of the Monte Carlo simulation takes 283.4 seconds. This shows the superiority of the proposed approach in terms of computational efficiency.

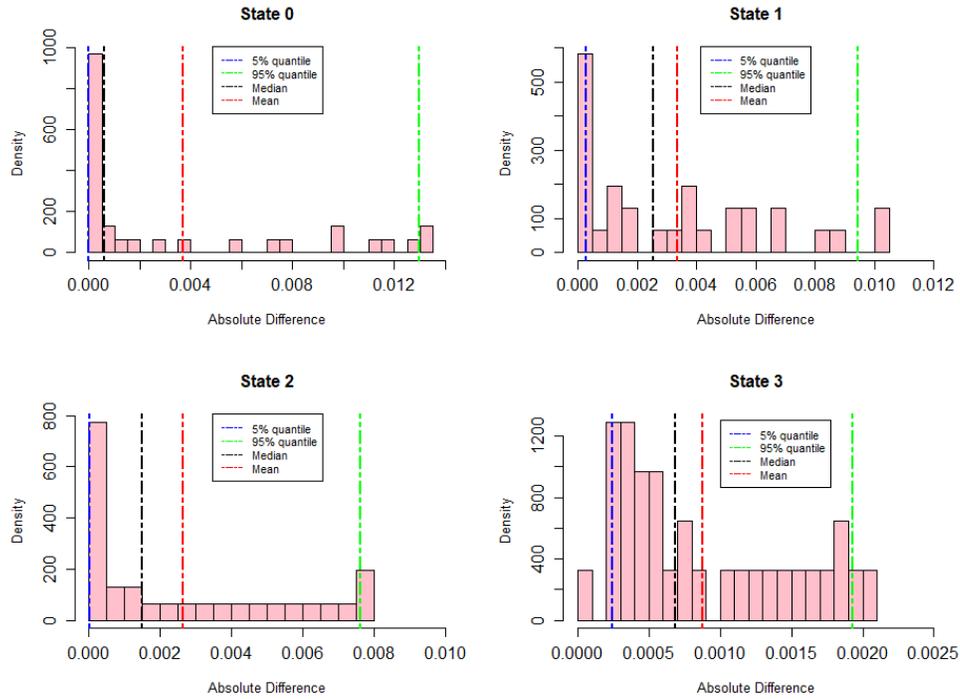

Figure 6. The absolute difference of the results between the proposed approach and the Monte Carlo simulation.

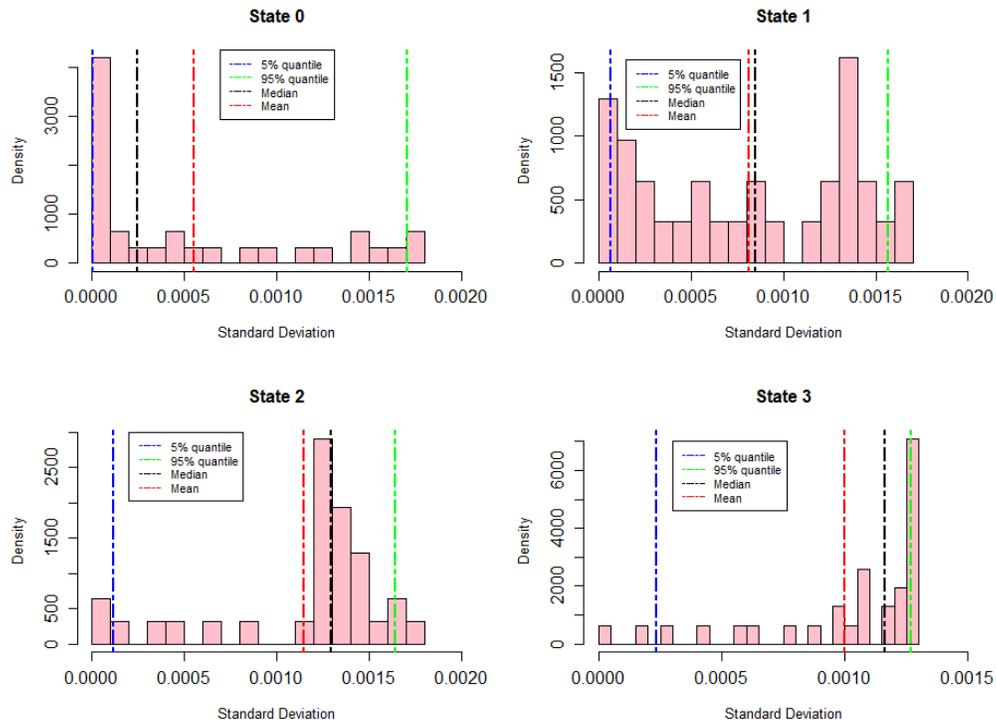

Figure 7. The composite standard deviation of the results between the proposed approach and the Monte Carlo simulation.

So far, we have demonstrated the performance of the proposed approach in assessing each system state probability. Then, we can calculate the system reliability by summing the probability in states 0 and 1. The results of system reliability are displayed in Figure 8 and show a good match for the results using all three methods. Therefore, we can conclude the validity of the proposed approach to assess system reliability.

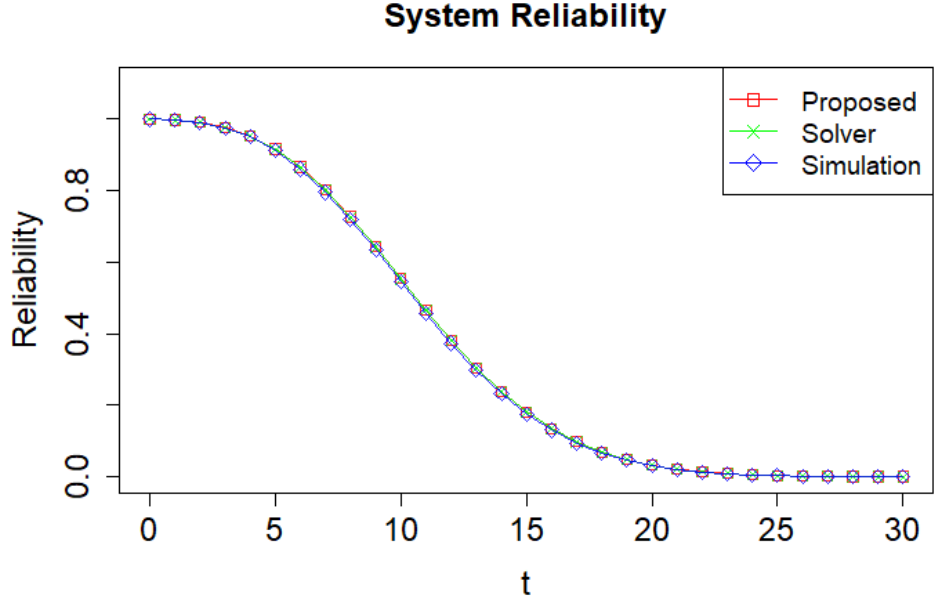

Figure 8. The results of system reliability using the proposed approach, the differential equation solver, and the Monte Carlo simulation.

*4.2.2. Example 2*

The proposed approach is of particular advantage for uncertainty quantification, which is important in reliability and safety applications. This example shows how the proposed approach can propagate and quantify the uncertainty of the system's initial condition. Denote the probability vector of the system's initial condition by $[\rho_0, 1 - \rho_0, 0,0,0,0]$, where $\rho_0$ follows the Beta distribution parameterized by two shape parameters, $\alpha = 5$ and $\beta = 1.5$. To integrate such uncertainty in the proposed approach, 50 samples are generated accordingly and are treated as a type of boundary condition, which needs to be satisfied by the neural network training process. In the generator, the neural network consists of 4 hidden layers of 50 neurons with the Tanh activation functions. There are four units in the final layer with the SoftMax activation function. The number of collocation points is 40, which are linearly spaced values generated within the range [0, 30]. In the discriminator, the neural network has 2 hidden layers of 50 neurons with the Tanh activation functions and 1 neuron in the output layer. An exponential decaying learning rate is applied with the starting learning rate as $1 \times 10^{-2}$, the decay rate as 0.9 and the decay step as 1000. The number of iterations is $1 \times 10^5$ using the Adam optimization algorithm. The weighting factor $\lambda$ is set equal to 1. Once the model is well trained, the generator is used to generate $5 \times 10^3$ samples to estimate the system state probability with uncertainty.

The Monte Carlo simulation consists of 50 replications and each replication includes $1 \times 10^5$ iterations. Note that each sample from the Monte Carlo simulation represents the actual state number, while the samples in the proposed approach represent the system state probability vector. Note that the Monte Carlo simulation is computationally expensive, which takes 300.6 seconds per replication and needs around 4 hours in total. However, the whole training and sampling process of the proposed approach takes only 625.9 seconds. This indicates the superior computational efficiency of the proposed approach that is 24 times more computationally efficient when compared with the Monte Carlo simulation.

Figures 9 and 10 display the predictions with uncertainty quantification for system state probability and system reliability, respectively. The results indicate the consistency between the proposed approach and the Monte Carlo simulation. Indeed, some deviations are observed in both the mean prediction and uncertainty bound for each state probability. Such deviation would be attributed to the sources of uncertainty due to the Monte Carlo simulation and the neural network configuration. This inconsistency would be further reduced by increasing the number of replications and iterations for the Monte Carlo simulation, enhancing the network configuration and training process. Note that the difficulty of training GANs has been well recognized, and the training of PIGANs becomes even more challenging due to the integration of more complicated composite generator loss. This is still an open topic on improving the configuration of network architecture and training of PIGANs, which is discussed in Section 5 and will be considered in the authors' future work.

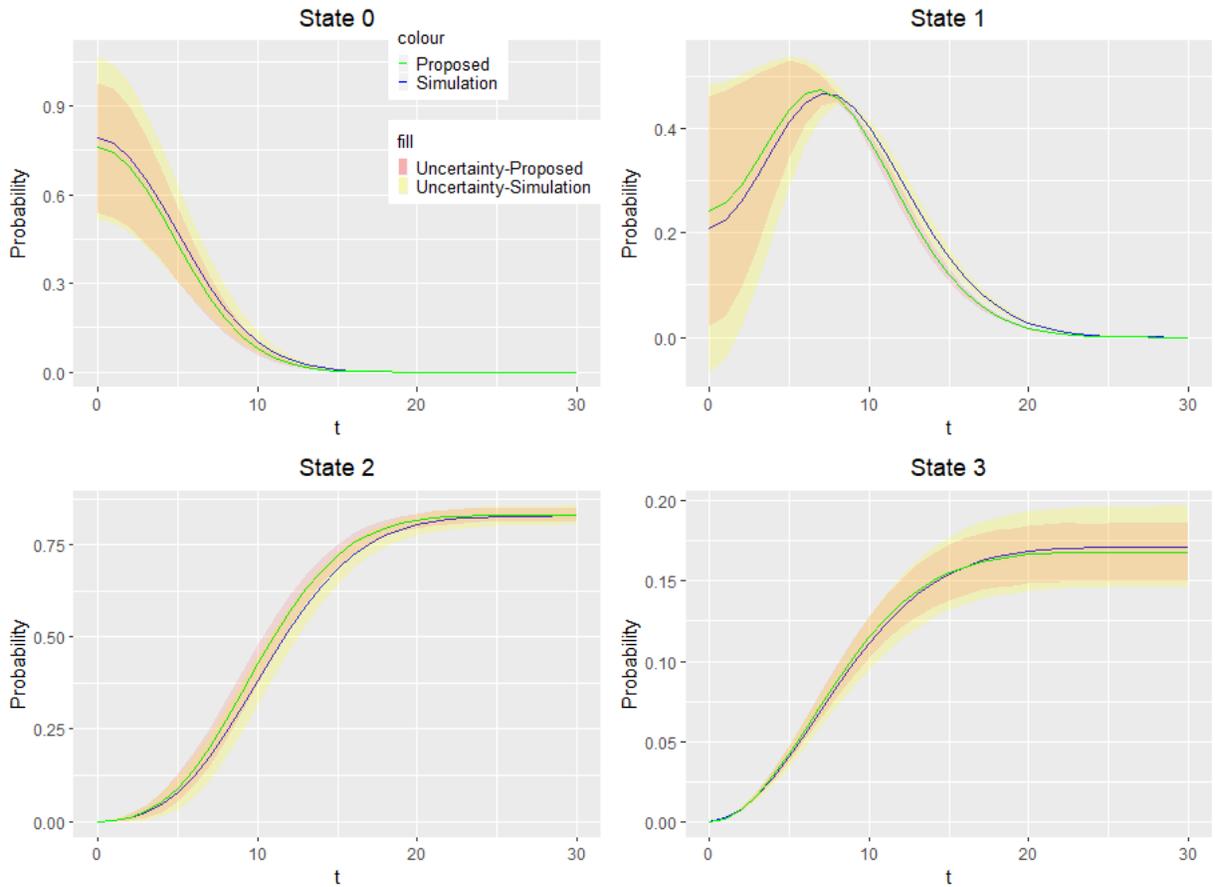

Figure 9. The results of system state probability using the proposed approach and the Monte Carlo simulation considering the measurement data of the system's initial condition.

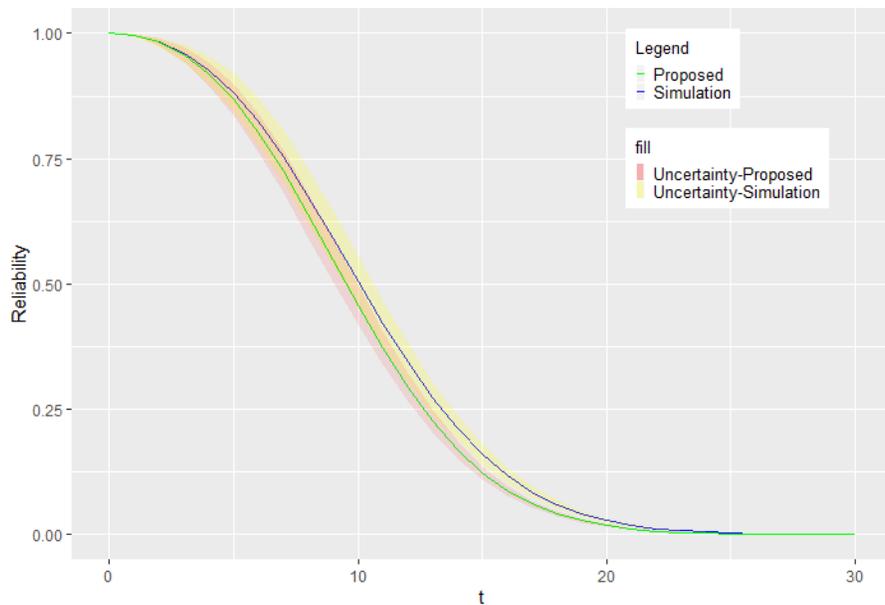

Figure 10. The results of system reliability using the proposed approach and the Monte Carlo simulation considering the measurement data of system initial condition.

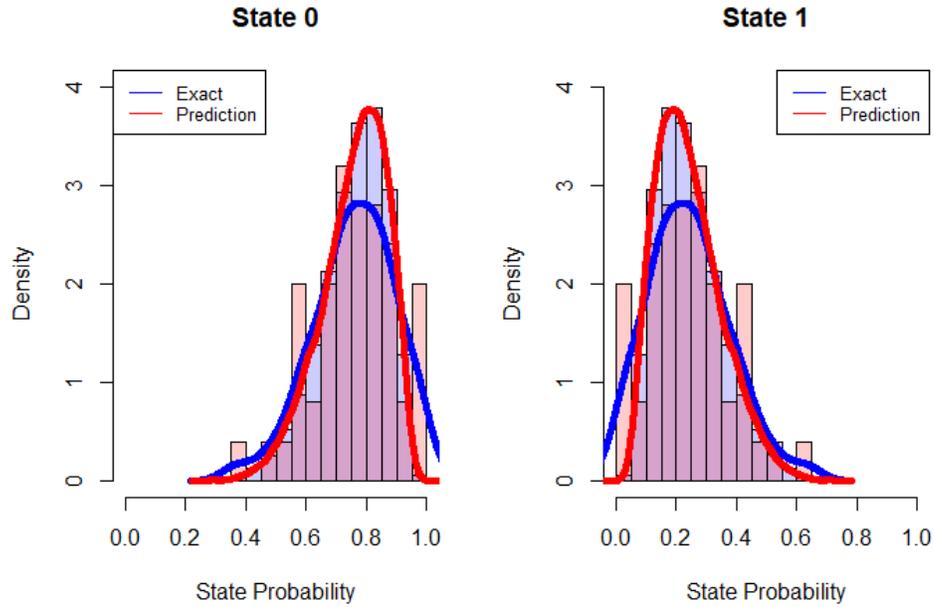

Figure 11. A comparison of the distribution of state probability initially in states 0 and 1 using the proposed approach.

We also present a comparison between the exact system's initial condition and the corresponding prediction by the proposed approach. A shown in Figure 11, the proposed approach performs well to quantify the uncertainty of the system's initial condition. Then, we follow the same process in Section 4.2.1 to evaluate the consistency of the results between the proposed approach and the Monte Carlo simulation. Figures 12 and 13 summarize the distribution of absolute difference and composite standard deviation between the proposed approach and the Monte Carlo simulation. As can be observed, both the two measures remain relatively small. Hence, the proposed approach provides a satisfactory result when compared with the Monte Carlo simulation.

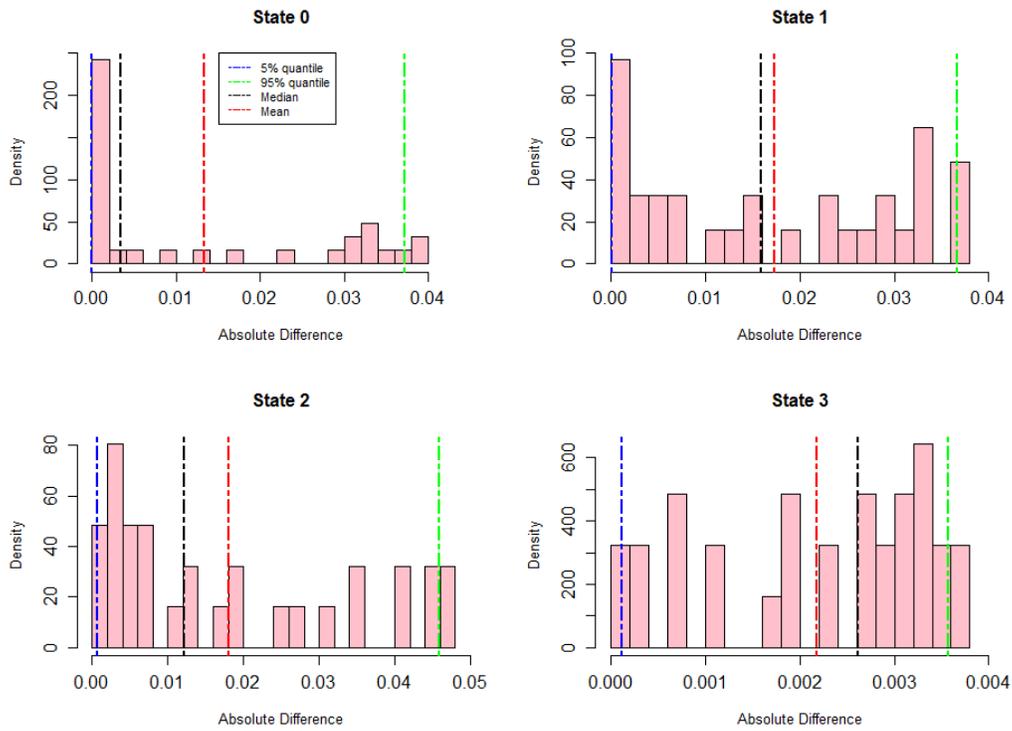

Figure 12. The absolute difference of the results between the proposed approach and the Monte Carlo simulation considering the measurement data of system initial condition.

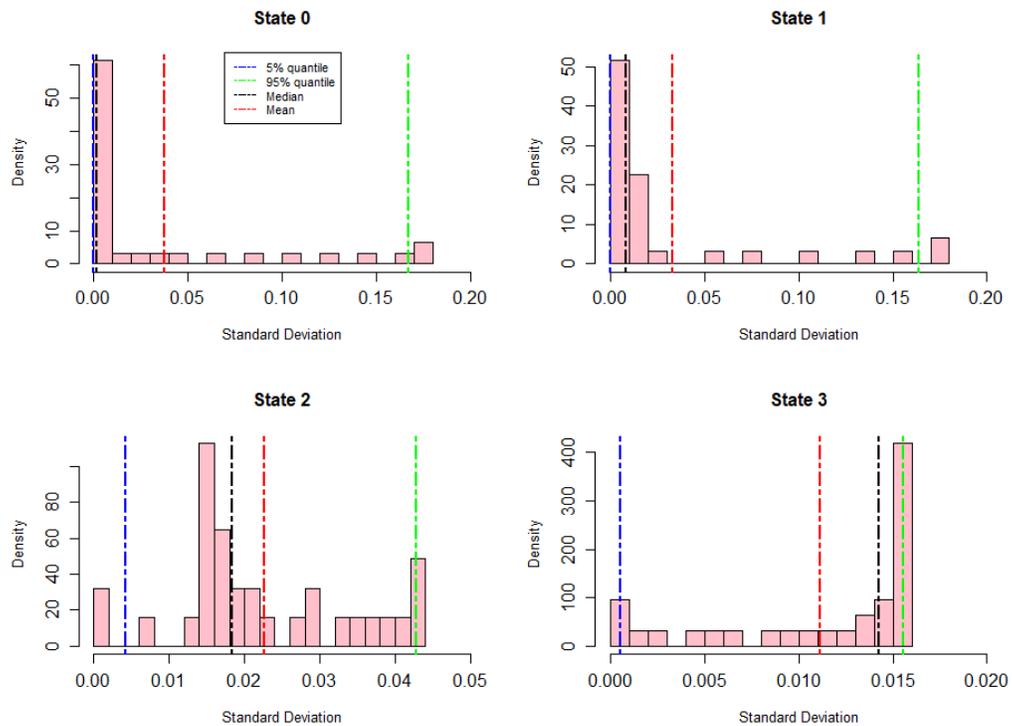

Figure 13. The composite standard deviation of the results between the proposed approach and the Monte Carlo simulation considering the measurement data of system initial condition.

*4.2.3. Example 3*

This example demonstrates the capability of the proposed approach to incorporate measurement data observed during a service life span. Note that the state-of-the-art method cannot incorporate this type of measurement data into system reliability assessment. Therefore, we heuristically demonstrate the process using a system with either better or worse performance when compared to the system in the first example as a baseline case. Specifically, synthetic measurement data are generated at various inspection time instances to reflect the performance deviation (i.e., better, or worse) from the baseline system. We incorporate these synthetic measurement data into system reliability assessment based on the proposed approach in Section 3.2 which, in turn, results in an updated evolution of system reliability up to mission time. Then one can heuristically evaluate the trend of reliability evolution in the simulated system. If the trend of updated reliability evolution is consistent with the assumption used to generate such synthetics measurement data, the proposed approach can be considered as effective to incorporate the measurement data collected during a service life span.

Suppose a simulated system is inspected at time $t$, and the corresponding measurement data is $[t, p_t]$, where $t$ is the simulated system's operational time, and $p_t$ is the state probability vector according to inspection and expert judgment. The same notation applies to the baseline case and the baseline system state probability is denoted by $[t^*, p_{t^*}^*]$. Then, the synthetic measurement data can be generated as equal to the state probability vector $p_t^*$ of the baseline case at a time instant $t^* = t \mp \Delta t$, which shifts the inspection time $t$ forward or backward. Particularly, a backward shift that is $t^* = t - \Delta t$ leads to a system with better performance; a forward shift that is $t^* = t + \Delta t$ leads to a system with worse performance.

Table 1: The synthetic measurement data generated to simulate a system with better or worse performance

|  | Inspection-Time $t$ | Time-Shifted $\Delta t$ | Synthetic Measurement Data | System Reliability |
|---|---|---|---|---|
| A system with better performance | 5 | 2 | [8.35E-01, 1.42E-01, 6.20E-03, 1.72E-02] | 0.977 |
|  | 10 | 2 | [2.79E-01, 4.47E-01, 1.81E-01, 9.23E-02] | 0.726 |
|  | 15 | 2 | [3.43E-02, 2.71E-01, 5.38E-01, 1.56E-01] | 0.3053 |
| A system with worse performance | 2 | 3 | [8.35E-01, 1.42E-01, 6.20E-03, 1.72E-02] | 0.977 |
| | 5 | 2 | [3.75E-01, 4.27E-01, 1.22E-01, 7.60E-02] | 0.802 |
| | 9 | 4 | [3.43E-02, 2.71E-01, 5.38E-01, 1.56E-01] | 0.3053 |

For demonstration purposes, Table 1 shows the synthetic measurement data generated to simulate two systems with either better or worse performance. The measurement data is sequentially used to update the system reliability evolution. The network architecture is used as the same as that is used in Section 4.2.2. The Adam optimization algorithm is used for training with $2 \times 10^4$ iterations. The weighting factor $\lambda$ is set equal to 1. An exponential decaying learning rate is applied with the starting learning rate as $1 \times 10^{-3}$, the decay rate as 0.9 and the decay step as 1000. Figures 14 and 15 show the reliability evolution of the simulated system with worse and better performance, respectively. The validity of results can be justified by the insights as below:

- The model can generally capture the trend that the reliability of the simulated system is almost always lower or larger than the baseline case as in Figures 14 and 15, respectively.
- The uncertainty of the reliability evolution can be effectively quantified to consider the measurement data since the synthetic measurement data are bounded by the two-standard deviation intervals of the updated system reliability.
- With more inspection data available in both simulated systems, the distance between the simulated system reliability curve and baseline system reliability curve becomes greater. This indicates that more measurement data make the model more confident to conclude that the simulated system is different from the baseline case.

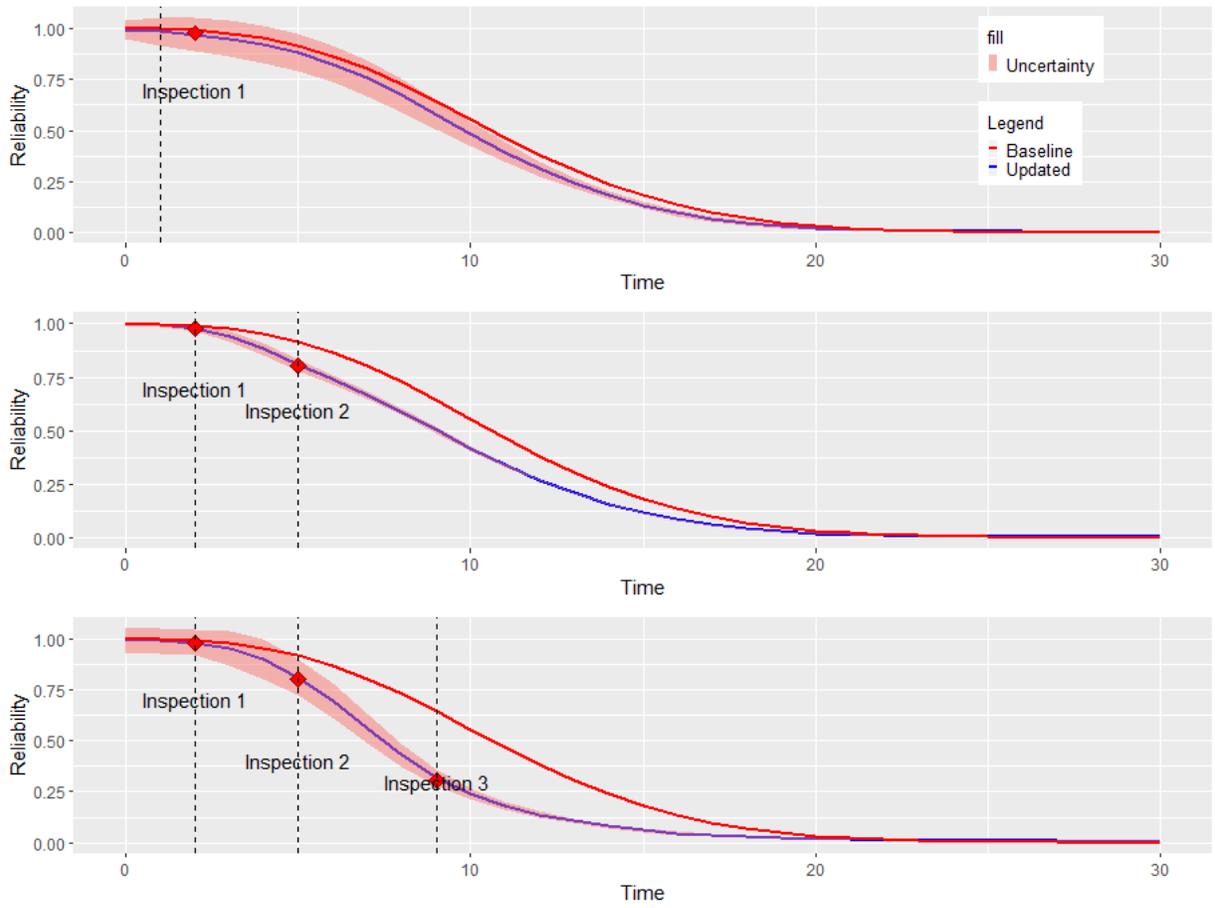

Figure 14. Updated reliability evolution for a simulated system with worse performance as compared to the baseline case.

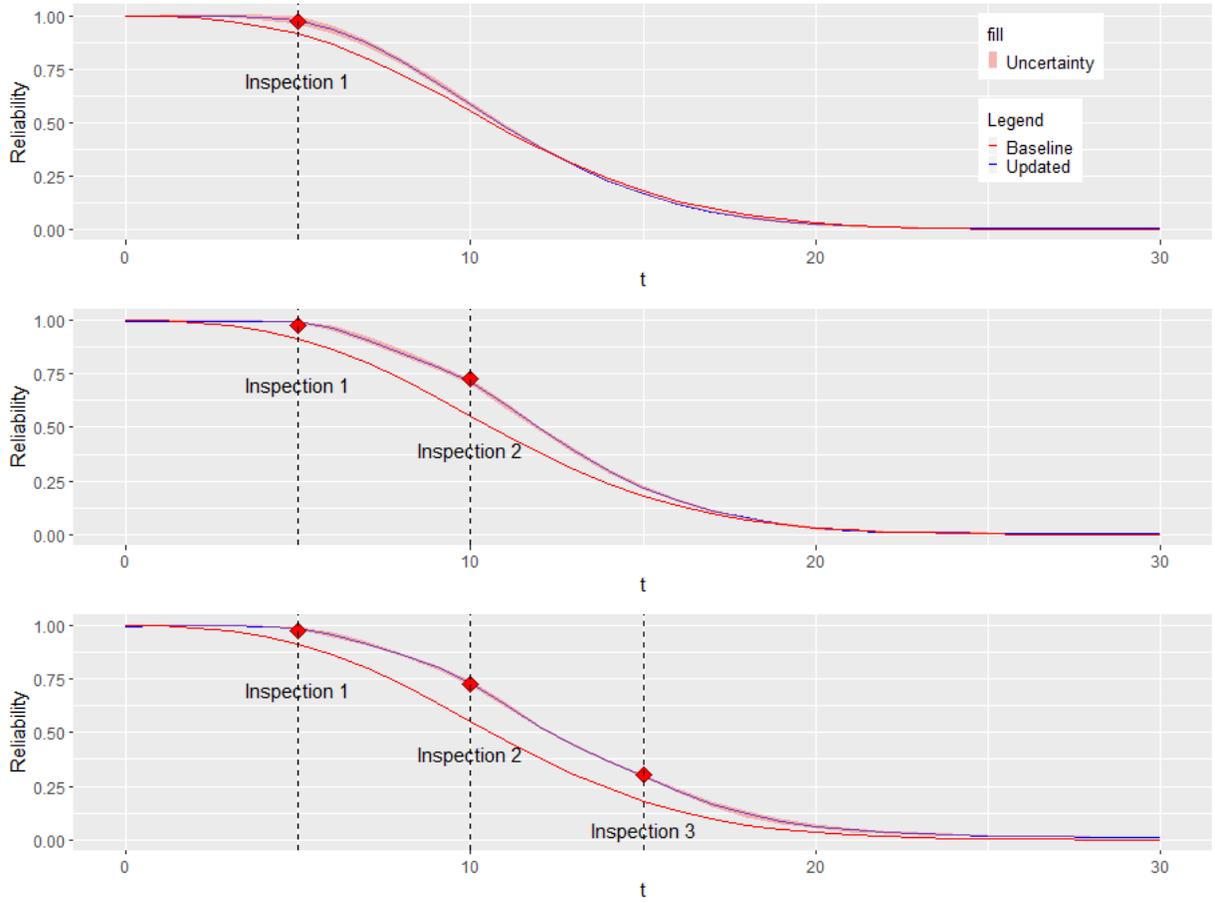

Figure 15. Updated reliability evolution for a simulated system with better performance as compared to the baseline case.

## 5. CONCLUSIONS AND FUTURE DIRECTIONS

This paper presented a novel perspective on system reliability assessment by leveraging the advance of physics-informed deep learning. We discussed the approach to formulate system reliability assessment in the deep learning context by encoding the system property into the network configuration and training which, in turn, led to a continuous solution to system reliability assessment that is parametrized by the neural network parameters. Those were accomplished based on the universal approximation theorem and automatic differentiation techniques. Upon bridging the gap between deep learning and system reliability assessment, we demonstrated the benefits of uncertainty quantification and measurement data incorporation using a PIGANs-based approach. The proposed approach was validated by three numerical examples using a dual-processor computing system. The model performance was examined by comparison with the differential equation solver and the Monte Carlo simulation. The results demonstrated that the physics-informed deep learning approach performs satisfactorily and has superior computational efficiency for system reliability assessment. Note that the results presented in this paper would be further improved by tuning the network configuration and training process. We hope that this paper can facilitate a better understanding and future exploration of the deep learning application to system reliability assessment, as we believe it holds many potential advantages for deep learning-based approaches for the reliability and safety community.

The systematic configuration of a robust network architecture and training algorithm for physics-informed deep learning is still an open and active research issue. Future work would be leveraging the advance of deep learning, such as use Bayesian deep learning and deep ensemble learning to enhance the capability of uncertainty quantification in system reliability assessment; exploit meta-learning and multi-task learning techniques to improve and facilitate the training process for system reliability assessment; develop new network configurations to address non-Markovian system reliability assessment; use physics-informed deep learning to discover new patterns of underlying system reliability evolution; develop a systematic framework to integrate measurement data and other mathematical models describing system reliability evolution.